\documentclass[conference,a4paper]{IEEEtran}
\IEEEoverridecommandlockouts
\usepackage{amsmath,amssymb,amsfonts}
\usepackage{booktabs}
\usepackage{graphicx}
\usepackage{multirow}
\usepackage{cite}
\usepackage{xcolor}
\usepackage{url}
\usepackage{xspace}
\usepackage{placeins}
\usepackage[hidelinks]{hyperref}
\usepackage[a4paper,total={184mm,239mm}]{geometry}

\newcommand{\method}{FlowPAC\xspace}

\begin{document}

\title{FlowPAC: Flow-Guided Visual Pruning and Action Caching for Efficient VLA Inference}

\author{%
\IEEEauthorblockN{%
Jintao Cheng\textsuperscript{1,$\dagger$},
Yipu Zhang\textsuperscript{1,$\dagger$},
Haozhe Wang\textsuperscript{2},\\
Lin Jiang\textsuperscript{3},
Jiang Xu\textsuperscript{4},
Wei Zhang\textsuperscript{1,*}}
\IEEEauthorblockA{%
\textsuperscript{1}The Hong Kong University of Science and Technology\\
\textsuperscript{2}South China Normal University\\
\textsuperscript{3}Northeastern University, China\\
\textsuperscript{4}The Hong Kong University of Science and Technology (Guangzhou)}
\thanks{$\dagger$ Equal contribution. * Corresponding author.}%
}

\maketitle

\begin{abstract}
Vision-Language-Action (VLA) models have emerged as a powerful paradigm for embodied AI, yet their high inference latency and memory footprint remain major barriers to edge deployment. Existing training-free acceleration techniques, such as visual token pruning and action caching, typically rely on static resource allocation. This mismatch with dynamic robotic interactions not only degrades control performance but also causes severe hardware inefficiencies, including fragmented KV-cache layouts and wasted compute on non-critical regions. To address this, we propose \textbf{FlowPAC}, a locality-aware framework that leverages task-aware optical flow as a shared physical signal to jointly optimize visual and action computation. Specifically, we introduce \textbf{Direction-Aware Trajectory Propagation (DATP)} and \textbf{Locality-Aware Contiguous Token Selection (LACT)} to jointly handle visual token dynamics and enforce spatial locality, reducing memory fragmentation and enabling efficient KV-cache compression. This is complemented by \textbf{Flow-Conditioned Action-Block Refresh (FCAR)} for adaptive action computation based on motion dynamics. Extensive closed-loop evaluations on LIBERO demonstrate that FlowPAC achieves a \textbf{2.13$\times$} end-to-end speedup over the dense baseline while \textbf{essentially maintaining the success rate} with 30\% visual-token retention. These results validate that aligning computation allocation with both physical dynamics and hardware constraints is essential for efficient VLA deployment on edge systems.
\end{abstract}


\section{Introduction}
Vision-Language-Action (VLA) models have emerged as a promising solution to enhance generalization of robotic control in embodied intelligence~\cite{brohan2022rt,brohan2023rt,team2024octo,kim2024openvla}. The mainstream VLA architecture pairs a pretrained VLM backbone with a dedicated action expert to unify visual, textual, and action information within a token-domain framework~\cite{black2024pi_0,intelligence2025pi05}. While implementations like the $\pi$ series demonstrate strong generalization, they suffer from prohibitive end-to-end latency on device deployment.

Recent work on model efficiency spans quantization of VLA policies~\cite{zhang2026quantvla} and visual geometry transformers~\cite{zhang2026fgq}, layer skipping~\cite{zhang2026mole}, knowledge distillation~\cite{dong2025vitavla}, visual token pruning~\cite{cheng2026vla}, and action-head feature caching~\cite{yang2026efficientvla}. Efficient multimodal inference has also been explored in robotic perception, including training-free visual place recognition~\cite{cheng2025scale}. Among these, visual token pruning and action caching have emerged as effective techniques for improving VLA inference efficiency, as they offer the distinct advantage of significantly reducing computational and memory overheads without requiring costly model retraining or architectural modifications~\cite{liu2025vla,xu2026vla,oi2026training}. Visual token pruning reduces spatial redundancy by scoring patches with instruction-conditioned attention, action-decoding relevance, or recent action dynamics and retaining a compact token set under a prescribed budget, thereby reducing subsequent token processing and KV-cache storage~\cite{pei2026action,liu2025vla}. Caching-based methods reduce temporal redundancy either by reusing visual-token KV states across adjacent observations~\cite{xu2026vla} or by retrieving intermediate action trajectories from similar multimodal contexts to warm-start iterative action generation~\cite{oi2026training}. However, the holistic impact of these disjointed optimizations on the end-to-end VLA pipeline remains under-explored.

\begin{figure*}[t]
    \centering
    \includegraphics[width=0.9\textwidth]{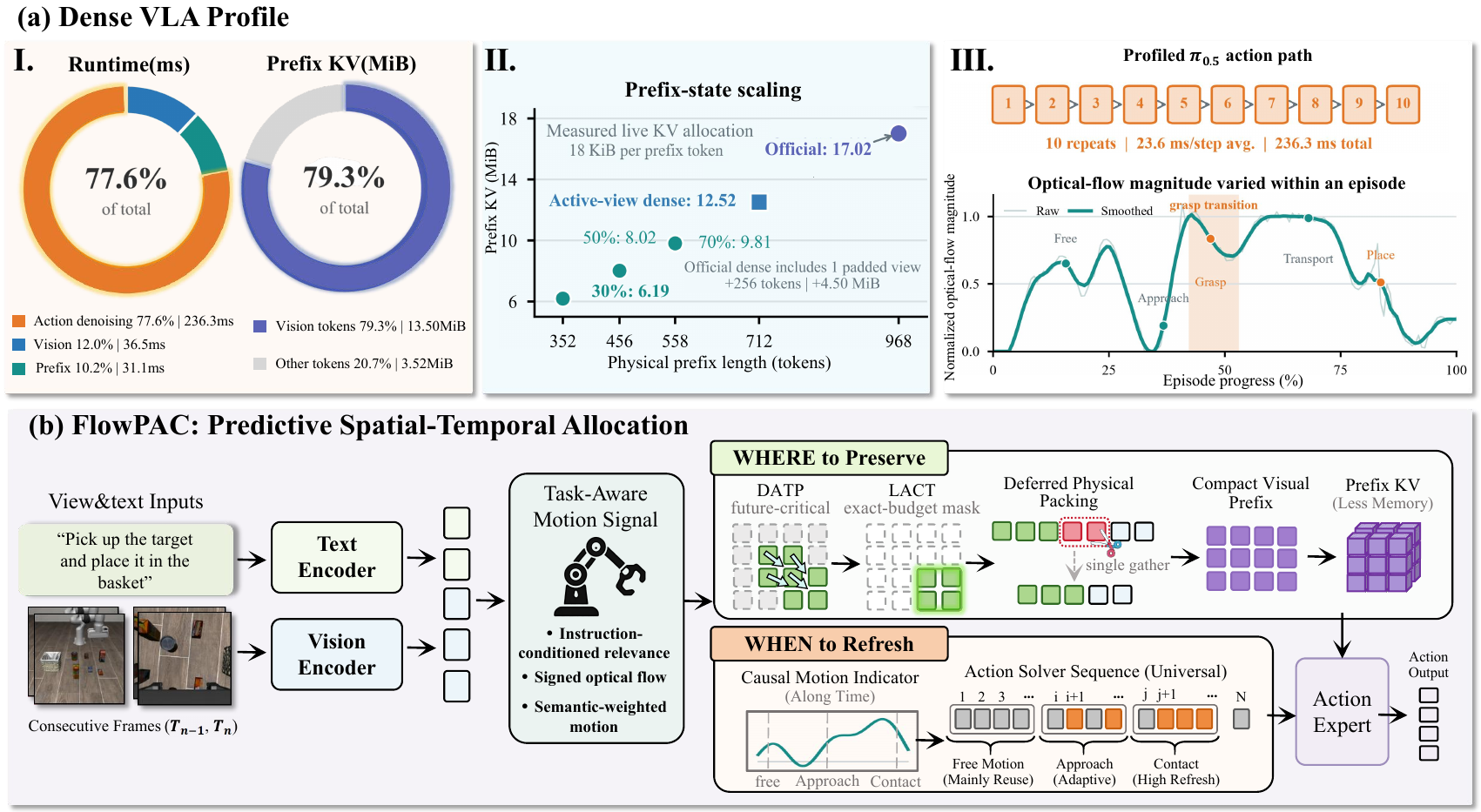}
    \caption{Profiling and overview of \method. (a) Dense $\pi_{0.5}$ profiling exposes two resource imbalances:
    action denoising dominates runtime (77.6\%), while visual tokens dominate the prefix KV state (79.3\%). The rightmost profiling panel shows optical-flow magnitude between consecutive observations, normalized within one representative rollout to visualize relative changes across task phases. (b) \method uses instruction-conditioned optical flow as a shared signal to coordinate \emph{where} visual tokens are preserved (DATP, LACT) and \emph{when} action-expert computation is refreshed (FCAR).}
    \label{fig:overview}
\end{figure*}



Crucially, adapting these isolated optimizations to resource-constrained robotic platforms is non-trivial~\cite{kwon2023mobile,lee2024vit,fan2026sata}. First, existing pruning methods typically decide which visual tokens to keep based on their importance in the current frame~\cite{pei2026action,liu2025vla}. However, a region that is unimportant in the current frame may become critical in the next few frames, for example, when the gripper moves toward it or approaches a contact boundary. Pruning decisions must be made before this future importance becomes directly observable. Premature pruning can therefore discard future-critical information and misallocate the compute budget, degrading control accuracy. Second, existing action-side caching methods typically use a fixed refresh rule, such as recomputing the action expert every few steps or when a predefined threshold is reached~\cite{chen2026real,oi2026training}. Yet recomputation needs vary throughout an interaction: frequent refreshes during smooth phases waste computation by repeating similar work, whereas infrequent refreshes during rapid or contact-rich transitions can reuse outdated action updates and introduce control lag. Fixed refresh rules thus fail to adapt computation to changing interaction dynamics. Both problems arise from the same limitation: the computational value of visual tokens and the freshness of cached action updates change with the interaction dynamics. This motivates a shared dynamic signal that can capture both where visual information is likely to become important and when cached action updates need to be refreshed.

To tackle these bottlenecks, we present \textbf{FlowPAC}, a training-free framework that leverages task-aware optical flow as a \textbf{shared physical signal} to restore the temporal and spatial continuity overlooked by existing acceleration methods. We design this framework to directly address the aforementioned bottlenecks. First, to mitigate the \emph{compute budget misallocation} caused by current-frame-only pruning, we introduce \textbf{Direction-Aware Trajectory Propagation (DATP)}. DATP exploits the fact that optical flow is a pixel-wise temporal derivative to propagate token importance along the signed flow field, explicitly preserving tokens in \textit{future} interaction regions regardless of their current redundancy. Second, we introduce \textbf{Flow-Conditioned Action-Block Refresh (FCAR)}, which adapts cache refreshes to optical-flow dynamics, reducing updates during stable phases and increasing them during transients to match computation with task demands. Furthermore, to ensure these optimizations yield genuine end-to-end speedups on resource-constrained robotic platforms instead of being negated by fragmented masking, we incorporate \textbf{Locality-Aware Contiguous Token Selection (LACT)}. By leveraging the inherent spatial locality of optical flow, LACT enforces contiguous block selection after a dense warm-up phase, reducing memory fragmentation while preserving physically meaningful token groups, thereby making the above dynamic scheduling practically deployable.


To summarize, this work makes the following contributions:
\begin{itemize}
    \item We propose DATP, which combines instruction-conditioned semantic relevance with signed optical flow to propagate semantic relevance toward future interaction regions and allocate visual tokens along predicted trajectories.

    \item We introduce FCAR, which maps semantic-weighted dual-view motion to a causal, episode-calibrated refresh budget. FCAR selectively reuses action-expert residuals during stable solver steps without shortening the original flow-matching trajectory.

    \item We develop LACT with deferred physical packing. LACT constructs an exact-budget, spatially coherent token mask, while deferred packing preserves early cross-token interaction and physically shortens the position-preserving visual prefix and its KV cache for subsequent layers.

\end{itemize}

\section{Profiling and Motivation}
\label{sec:profiling_motivation}    

As illustrated in the introduction, VLA systems typically consist of a VLM backbone and an action expert. To gain deeper insights into their execution efficiency, we profile $\pi_{0.5}$ as our representative model. 


\subsection{Profiling Setup}
\label{subsec:profiling_setup}

To identify the computational and memory bottlenecks of flow-matching VLA inference, we profile the complete $\pi_{0.5}$ policy inference under the same closed-loop rollout protocol used for evaluation. Our analysis targets three questions: (1) where execution time concentrates, (2) how the visual-prefix KV state scales with token count, and (3) whether optical-flow magnitude between consecutive observations varies across task phases. All measurements are taken on a single NVIDIA A100-PCIE-40GB GPU with BF16 precision and batch size one. 
The live tensor allocation of the visual-prefix KV state is captured at each retention setting. To construct the optical-flow magnitude trace, we compute dense optical flow between each pair of consecutive rollout frames and average the magnitudes of the top 15\% of pixels within each frame pair. Here, the magnitude of a flow vector denotes its Euclidean norm, consistent with Sec.~\ref{subsec:fcar}. We show the raw trace together with an offline centered moving-average trace to suppress frame-level noise and apply within-rollout percentile normalization for visualization.

\subsection{Bottleneck Analysis}
\label{subsec:motivations}

Panel~\textbf{I} of Fig.~\ref{fig:overview}(a) reveals two distinct resource imbalances between the sources of latency and persistent memory in dense $\pi_{0.5}$ inference: action denoising dominates runtime (77.6\%), while visual tokens dominate the prefix KV state (79.3\%).

As shown in panel~\textbf{II} of Fig.~\ref{fig:overview}(a), visual tokens account for 768 of the 968 prefix tokens (79.3\%) and occupy 13.50 of the 17.02\,MiB live KV state. Across the evaluated retention settings, physical packing reduces the prefix KV state from 17.02\,MiB to 9.81, 8.02, and 6.19\,MiB, showing that physically shortening the prefix substantially reduces persistent KV storage. However, vision encoding and prefix computation together account for only 22.2\% of policy-core latency, so even their idealized removal yields a visual-only speedup ceiling of $1/0.778 = 1.29\times$. Visual pruning alone is therefore insufficient; repeated action denoising, the dominant latency component, must also be addressed.


At each control query, the $\pi_{0.5}$ action expert performs ten flow-matching denoising steps, consuming 236.3\,ms, or 77.6\% of the 304.6\,ms policy-core latency. However, the action expert does not need to be recomputed at every step. When the interaction changes slowly, consecutive observations are similar, so the previous action update can be reused. During more dynamic interactions, such as grasping or contact, the scene changes more quickly and the cached update may no longer be accurate, requiring more frequent recomputation. Therefore, the key is to decide when the cached update can be reused and when it should be refreshed. As shown in the rightmost panel (III) of Fig.~\ref{fig:overview}(a), the optical-flow magnitude also changes across different interaction phases, with larger values during more active interactions. This suggests that optical flow can provide a simple observable signal for deciding when to reuse the cached update and when to recompute the action expert.

\section{Method}

\subsection{Overview}
\label{sec:overview}
\method jointly optimizes visual and action computation by using task-aware optical flow as a shared physical signal to coordinate two complementary axes: \emph{where} visual computation is preserved and \emph{when} action-expert computation is refreshed.
As summarized in Fig.~\ref{fig:overview}(b), the spatial branch comprises DATP, which predicts future-critical tokens using instruction-conditioned relevance and signed motion, and LACT, which converts the spatial score into an exact-budget, spatially coherent mask followed by deferred physical packing.
The temporal branch, FCAR, aggregates the same task-aware motion evidence into a causal refresh signal that adapts action-expert recomputation to the current dynamics of the interaction.

Let $I_t^v$ denote the observation from view $v\in\{g,w\}$ at control query $t$, where $g$ and $w$ denote the global and wrist camera views, respectively, and let $x$ be the language instruction.
At each query, \method constructs per-view task-aware motion evidence from instruction-conditioned visual relevance and signed optical flow between consecutive observations.
This shared signal drives both the spatial and temporal branches described below.

\subsection{Direction-Aware Trajectory Propagation}

\begin{figure}[!t]
    \centering
    \includegraphics[width=\columnwidth]{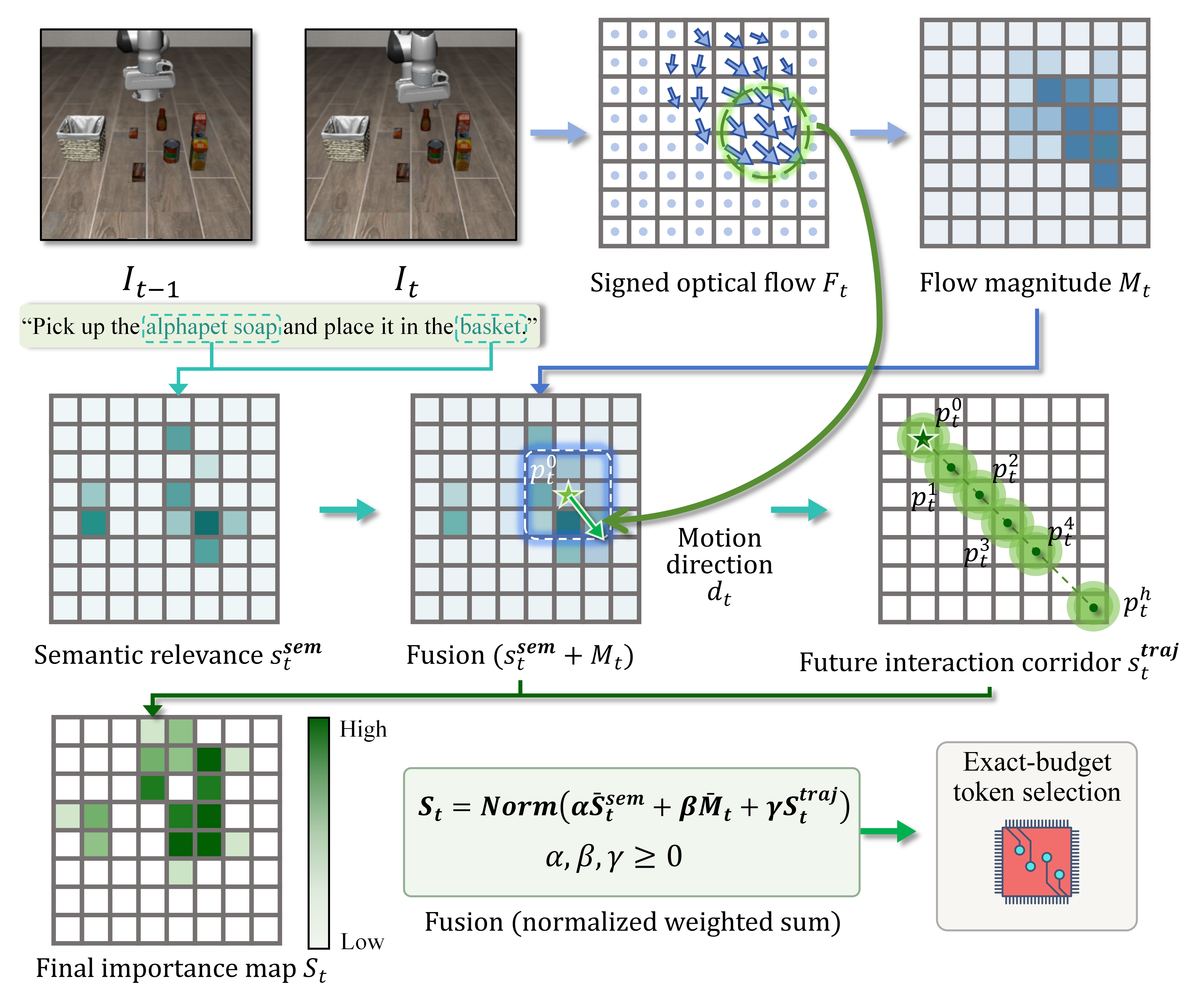}
    \caption{Direction-aware trajectory propagation. DATP advects instruction-conditioned semantic relevance along signed optical flow over multiple future horizons and merges them into a future-critical interaction corridor.}
    \label{fig:datp}
\end{figure}

As illustrated in Fig.~\ref{fig:datp}, DATP addresses a key limitation of current-frame token selection: patches that appear redundant now may become important shortly after due to object or camera motion.
Rather than selecting tokens solely based on their importance in the current frame, DATP extends the current task-relevant regions along their observed motion direction, forming a short-horizon spatial prior that preserves regions likely to become important in future interaction.

We first compute dense optical flow between consecutive observations $I_{t-1}^{v}$ and $I_{t}^{v}$, then aggregate pixel-wise flow to patch-level motion vectors $F_{t,i}^{v}$, preserving both magnitude and direction.
We then identify task-relevant regions using the instruction-conditioned relevance score $s_{t,i}^{\mathrm{sem},v}$ and estimate their collective motion center $p_t^{0,v}$ and dominant direction $d_t^v$ by semantic-weighted pooling of the flow field.
Starting from this center, DATP propagates along the motion direction for $H$ steps:
\begin{equation}
p_t^{h,v}
=
p_t^{0,v}
+
h\Delta d_t^v,
\qquad
1 \leq h \leq H,
\label{eq:trajectory-propagation}
\end{equation}
where $\Delta$ controls the spatial step size. If the propagated locations move outside the image, we limit them to the nearest image boundary. If the task-conditioned motion signal is too weak, DATP does not use the spatial prior for the current observation to avoid introducing an unreliable spatial bias.

Each visual patch then receives a trajectory score based on its proximity to the propagated path:
\begin{equation}
s_{t,i}^{\mathrm{traj},v}
=
\max_{1 \leq h \leq H}
\left[
\kappa^{h-1}
\exp
\left(
-\frac{
\left\|
p_i-p_t^{h,v}
\right\|_2^2
}{
2\sigma^2
}
\right)
\right],
\label{eq:trajectory-relevance}
\end{equation}
where $p_i$ is the normalized center of patch $i$, $\kappa \in (0,1]$ discounts distant future locations, and $\sigma$ controls the corridor width.
The $\max$ operation allows a patch to receive high relevance if it lies near \emph{any} propagated location.

Finally, we combine instantaneous relevance, motion magnitude, and future spatial relevance into the token-selection score:
\begin{equation}
s_{t,i}^{v}
=
\mathrm{Norm}
\left(
\alpha
\mathrm{Norm}
\left(
s_{t,i}^{\mathrm{sem},v}
\right)
+
\beta
\mathrm{Norm}
\left(
m_{t,i}^{v}
\right)
+
\gamma
s_{t,i}^{\mathrm{traj},v}
\right).
\label{eq:datp-score}
\end{equation}
Tokens are ranked independently per view by $s_{t,i}^{v}$ and allocated to the fixed budget $K_v$.

Overall, DATP extends token importance estimation from a purely current-frame perspective to a short-horizon spatial perspective, without introducing additional training or an explicit trajectory prediction model.

\subsection{Locality-aware Contiguous Token Selection}

\begin{figure}[!t]
    \centering
    \includegraphics[width=\columnwidth]{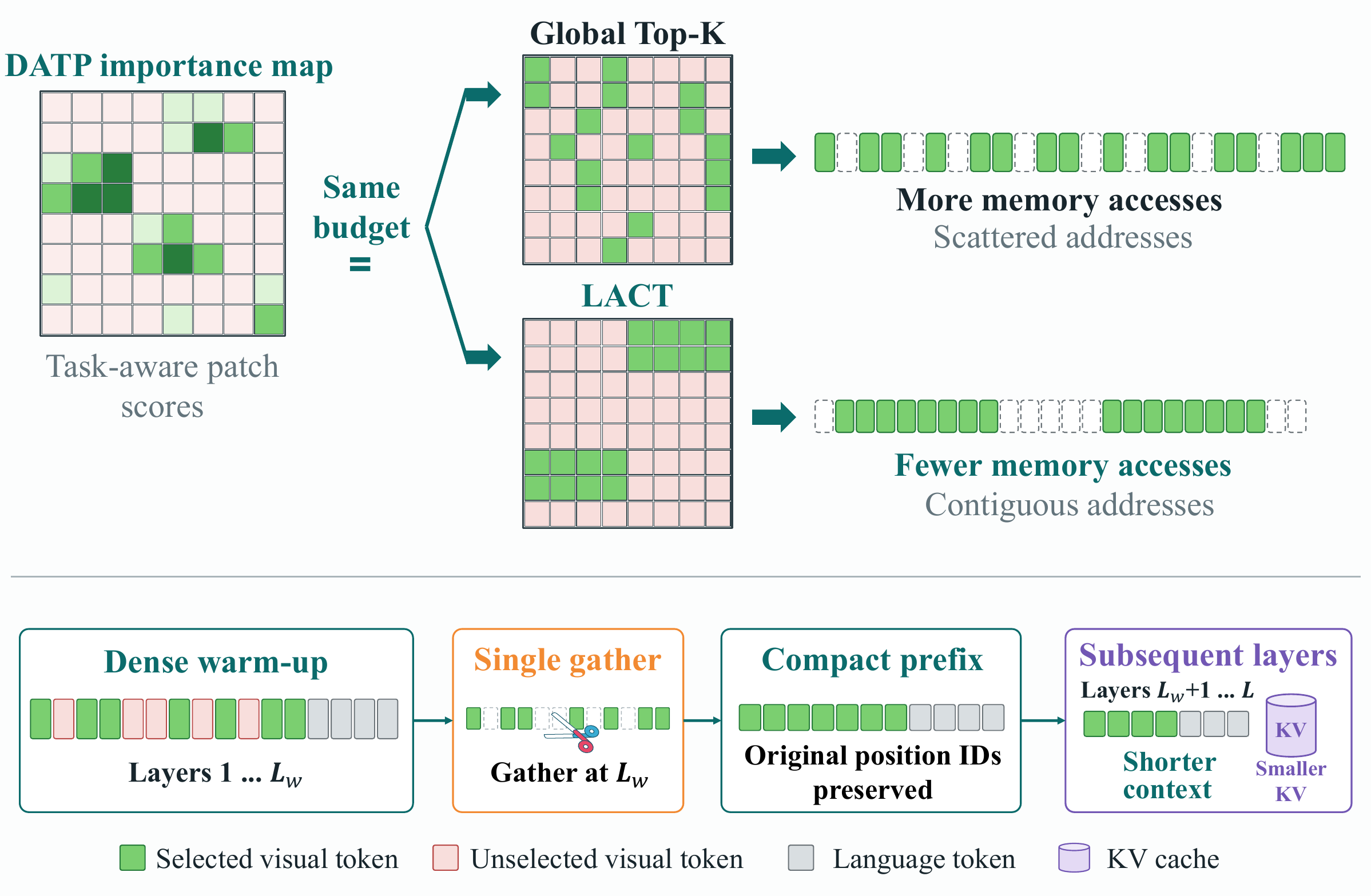}
    \caption{LACT and deferred physical packing. At the same exact budget $K$, LACT converts fragmented selections into fewer packing-gather memory accesses. A single intermediate gather preserves language tokens and original visual positions while shortening all subsequent prefix and KV computation.}
    \label{fig:hact}
\end{figure}

Common global Top-$K$ selection often produces spatially fragmented masks, where selected tokens are scattered across the image. Here, one memory access refers to reading one contiguous span of selected patches in raster order during the packing gather; fragmented masks therefore require many separate memory accesses, reducing the hardware benefit of token pruning. As shown in Fig.~\ref{fig:hact}, LACT groups visual tokens into aligned $b\times b$ blocks and searches over one- or two-component block templates around high-score trajectory regions. Complete blocks are selected first to preserve spatial coherence, while the remaining budget is filled by a partial block drawn from adjacent high-score patches. In this way, LACT replaces the scattered token selection of global Top-$K$ with the selection of spatially contiguous blocks. This makes the retained tokens easier to gather and physically compact, while the block budget is adjusted to keep exactly the same number of tokens as the original Top-$K$ selection. Conceptually, this procedure approximates the following structured selection objective:
\begin{equation}
\max_{m_t^v}
\quad
\sum_i s_{t,i}^v m_{t,i}^v
-
\lambda \mathcal{C}(m_t^v),
\qquad
\sum_i m_{t,i}^v = K_v,
\label{eq:hact-objective}
\end{equation}

where $m_t^v$ is the binary selection mask for view $v$, $K_v$ is the retained visual-token budget for that view, and $\mathcal{C}(\cdot)$ measures the spatial and memory fragmentation of the selected tokens.
LACT primarily enables downstream physical sequence shortening rather than optimizing memory access in isolation. Its contiguous block selection reduces the number of memory accesses required by the packing gather, while preserving spatially coherent interaction regions and producing a stable selection mask. This mask then allows a single gather operation to compact the retained tokens into a physically shorter prefix, directly reducing prefix length and KV-cache footprint.

However, directly packing the selected visual tokens at the input would disrupt early visual--language interaction and lead to substantial closed-loop performance degradation. We therefore defer physical pruning to a designated prefix layer $L_w$. The first $L_w$ layers operate on the original dense sequence, after which a single gather operation packs only the selected visual states. Language tokens remain unchanged and preserve their original order, while retained visual tokens keep their original rotary position IDs. The corresponding prefix KV tensors are compacted using the same index map.

Consequently, all subsequent prefix layers and action-expert cross-attention operations operate on a genuinely shorter context of length
\begin{equation}
N_t'
=
K_g + K_w + N_{\mathrm{lang}}
<
N_g + N_w + N_{\mathrm{lang}},
\label{eq:packed-length}
\end{equation}
where $N_t'$ denotes the packed prefix length at query $t$, $N_g$ and $N_w$ are the original global- and wrist-view token counts, $K_g$ and $K_w$ are the retained counts after LACT, and $N_{\mathrm{lang}}$ is the language-token count. 

\subsection{Flow-Conditioned Action-Block Refresh}
\label{subsec:fcar}
\begin{figure}[!t]
    \centering
    \includegraphics[width=\columnwidth]{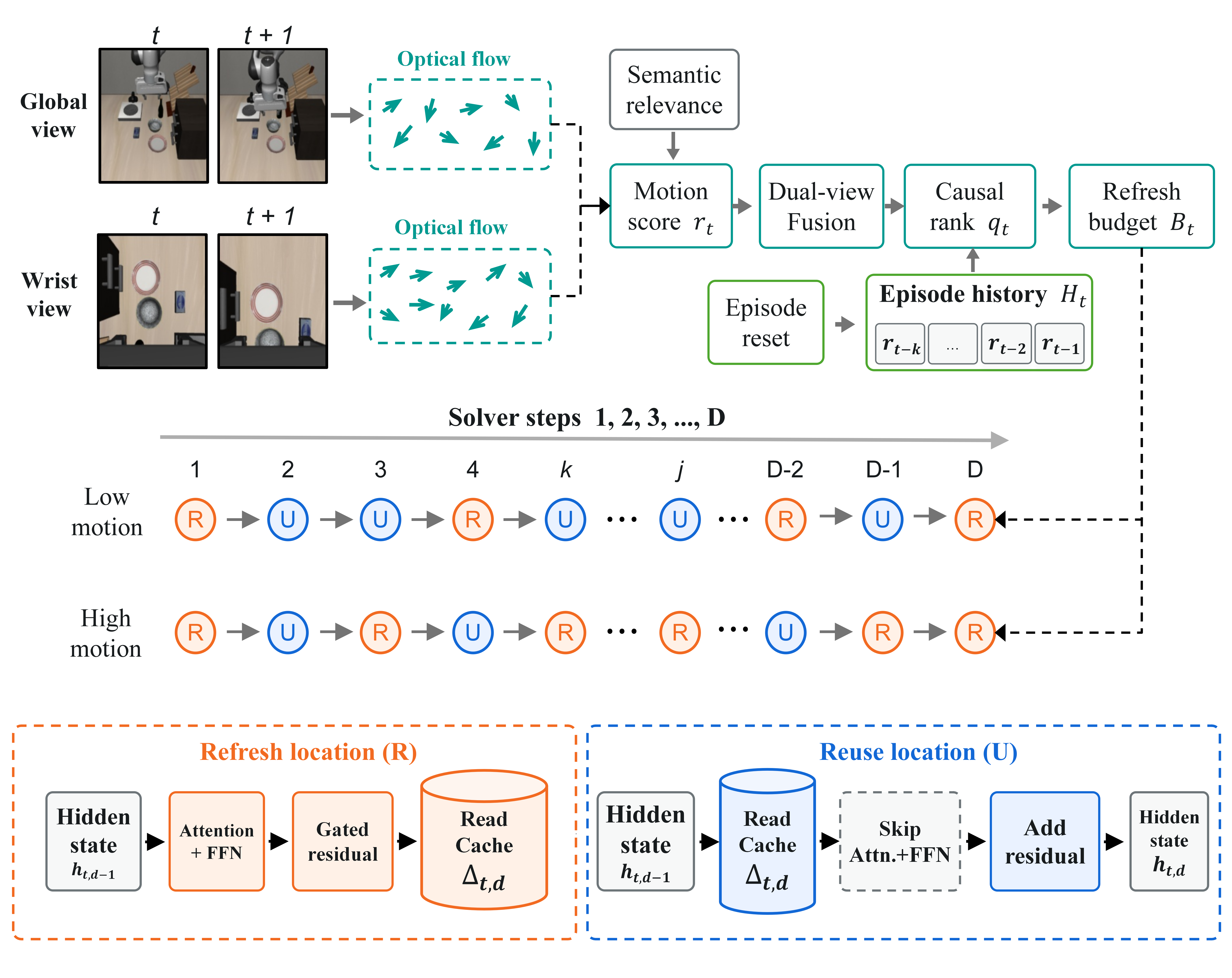}
    \caption{Flow-conditioned action-block refresh. Dual-view task-aware motion is calibrated by the causal episode history and converted into a refresh budget. Refresh steps compute and store attention/FFN residuals, whereas stable steps reuse them while preserving all action-state updates.}
    \label{fig:rdac}
\end{figure}

Physical visual packing reduces the prefix computation cost, but the action expert is still repeatedly evaluated across the flow-matching process. As shown in Fig.~\ref{fig:rdac}, FCAR addresses this redundancy by converting task-aware visual dynamics into an episode-adaptive recomputation budget. For each view, we measure the semantic-weighted motion magnitude as
\begin{equation}
r_t^v
=
\sum_i
\hat{s}_{t,i}^{\mathrm{sem},v}
\left\|
F_{t,i}^v
\right\|_2,
\qquad
r_t=\max(r_t^g,r_t^w),
\label{eq:motion-risk}
\end{equation}
where $\hat{s}_{t,i}^{\mathrm{sem},v}$ is normalized to sum to one. Taking the maximum across the global and wrist views makes the gate conservative: significant motion in either view is sufficient to trigger additional computation.

Rather than relying on a hand-tuned task-dependent threshold, FCAR converts $r_t$ into a causal empirical percentile within the current episode:
\begin{equation}
q_t
=
\frac{
1+\sum_{\tau<t}
\mathbf{1}\{r_\tau\leq r_t\}
}{
t+1
}.
\label{eq:causal-percentile}
\end{equation}
The first query is assigned $q_0=1$, and the motion history is cleared whenever a new episode begins. Given the minimum and maximum refresh fractions $(\rho_{\min},\rho_{\max})$, the number of action-expert evaluations retained at step $t$ is determined by
\begin{equation}
B_t
=
\operatorname{clip}
\left(
\operatorname{round}
\left[
D
\left(
\rho_{\min}
+
q_t(\rho_{\max}-\rho_{\min})
\right)
\right],
2,D
\right),
\label{eq:refresh-budget}
\end{equation}
where $D$ denotes the original number of flow-matching steps ($D=10$ in our $\pi_{0.5}$ evaluation). The $B_t$ refresh locations are distributed uniformly along the trajectory, including its two boundaries.

At a refresh location, each action-expert layer performs its attention and feed-forward network (FFN) operations normally and caches the resulting residual updates. At a reuse location, both operators are skipped, and the most recently cached residuals are added to the current hidden state. Importantly, FCAR does not alter the flow-matching solver or reduce its integration steps: the action state and timestep embedding are updated at all $D$ solver steps. FCAR therefore reduces computation by selectively reusing action-expert block evaluations while preserving the original solver trajectory.

As a result, low task-conditioned motion leads to more aggressive reuse, whereas significant motion in either the global or wrist view automatically increases the refresh budget. This enables FCAR to allocate computation according to the instantaneous dynamics of the interaction rather than using a fixed temporal caching schedule.

\subsection{Joint Inference Procedure}
For every control query, DATP computes semantic and motion evidence once and exposes it to both branches. LACT uses the spatial field to physically pack a future-aware visual prefix, while FCAR reduces action-expert recomputation according to the scalar dual-view motion risk. The modules are therefore coupled by a shared physical signal but optimize different axes: token packing reduces prefix length and KV storage, and action refresh reduces repeated temporal computation. 

\section{Evaluation}

\subsection{Experimental Setup}
\label{sec:experimental-setup}
We evaluate on the official $\pi_{0.5}$ LIBERO checkpoint~\cite{liu2023libero} with two active camera views and the original 10-step flow-matching solver.
LIBERO comprises four suites (Spatial, Object, Goal, Long) with 10 tasks each; every success-rate entry averages 50 trials per task (2{,}000 episodes total).
We compare against dense inference, FastV~\cite{chen2024image}, VLA-ADP~\cite{pei2026action}, and VLA-Cache~\cite{xu2026vla} at matched 70\%, 50\%, and 30\% control ratios.
For pruning methods (FastV, VLA-ADP), the ratio denotes visual-token retention; for VLA-Cache, the complementary non-reuse ratio.
All \method results share one fixed architecture (physical packing + FCAR) and differ only in retention.
All measurements run on one NVIDIA A100-PCIE-40GB with BF16 and batch size one, using synchronized CUDA events over the full inference path. We report median (P50) and 95th-percentile (P95) latencies, corresponding to the values below which 50\% and 95\% of runs fall, respectively; speedup is computed from P50.

\subsection{End-to-End Results}
\begin{figure}[t]
    \centering
    \includegraphics[width=0.9\columnwidth]{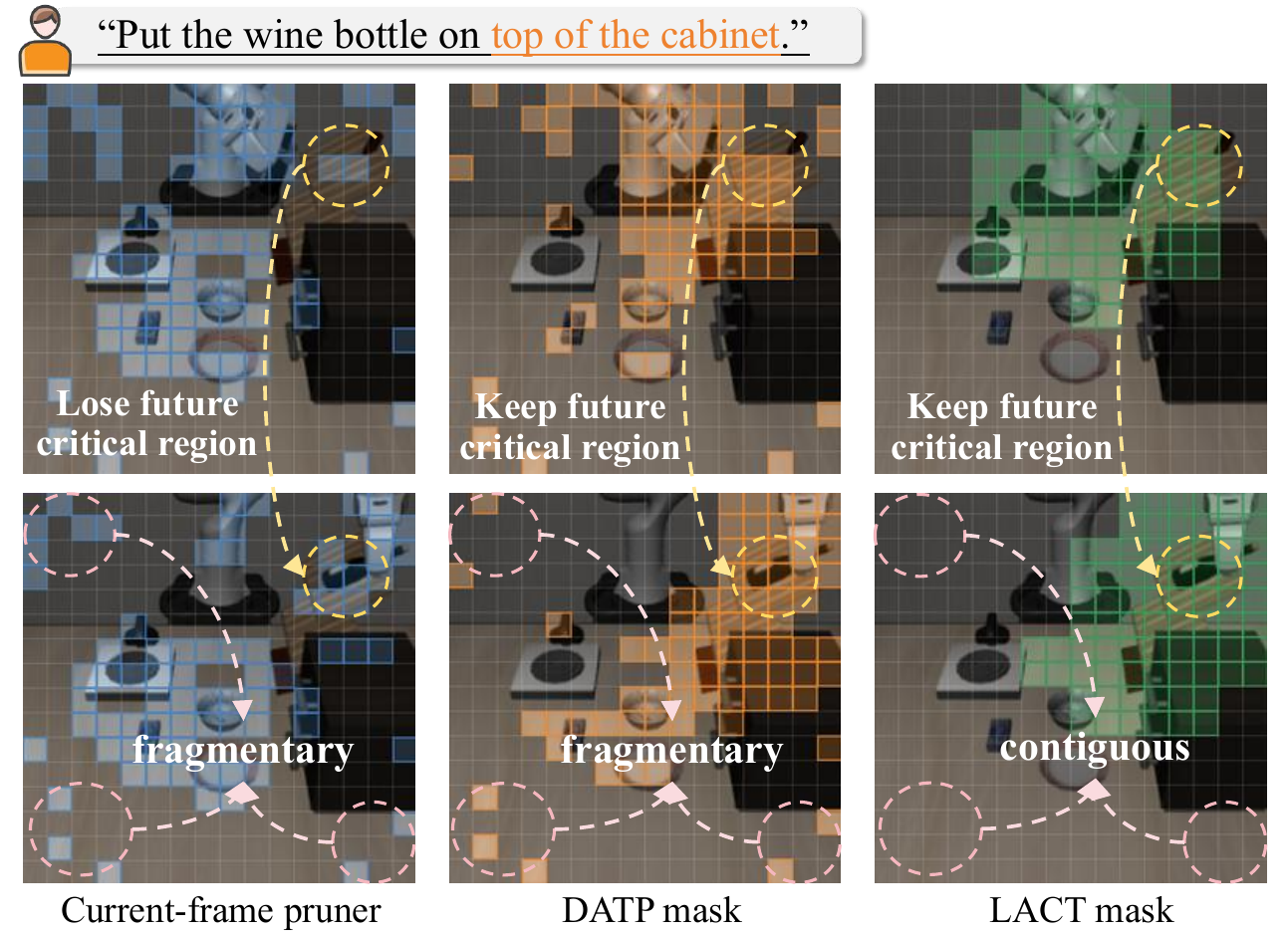}
    \caption{Qualitative comparison of visual token masks.}
    \label{fig:selection-visualization}
    \vspace{-0.6cm}
\end{figure}

Table~\ref{tab:success} reports the full success grid.
Current-frame pruning methods such as FastV and VLA-ADP cannot reliably identify and preserve regions that become relevant in subsequent interactions. Discarding these future-critical visual tokens deprives the policy of visual evidence needed for later action generation, reducing task success; this limitation is particularly evident on the long-horizon LIBERO-Long suite. Under the tight 30\% budget, FastV and VLA-ADP achieve only 71.2\% and 83.2\% success on LIBERO-Long, respectively, compared with 92.6\% for dense inference. VLA-Cache retains all visual tokens but reaches only 89.2\% on LIBERO-Long under aggressive visual-state reuse. In comparison, \method achieves 92.2\% on the same suite and 97.15\%, 96.15\%, and 95.80\% average success at 70\%, 50\%, and 30\% retention, respectively. Our flow-guided mask ranking preserves predicted future interaction regions while removing less task-relevant tokens, as qualitatively illustrated in Fig.~\ref{fig:selection-visualization}.

\begin{table}[t]
\caption{LIBERO success rate (\%). Each completed entry averages 10 tasks $\times$ 50 trials. \method uses one physical token-packing and FCAR architecture at three visual-token retention settings.}
\label{tab:success}
\centering
\scriptsize
\setlength{\tabcolsep}{8.1pt}
\begin{tabular}{lccccc}
\toprule
Method (ratio) & Spa. & Obj. & Goal & Long & Avg. \\
\midrule
Dense $\pi_{0.5}$ (100\%) & 98.4 & 98.0 & 95.6 & 92.6 & 96.15 \\
\midrule
FastV (70\%) & 97.4 & 99.4 & 97.6 & 93.0 & 96.85 \\
FastV (50\%) & 97.0 & 98.4 & 93.2 & 91.6 & 95.05 \\
FastV (30\%) & 86.4 & 96.6 & 87.2 & 71.2 & 85.35 \\
\midrule
VLA-ADP (70\%) & 96.6 & 98.8 & 96.8 & 90.8 & 95.75 \\
VLA-ADP (50\%) & 95.4 & 97.6 & 95.8 & 88.0 & 94.20 \\
VLA-ADP (30\%) & 93.4 & 97.4 & 95.8 & 83.2 & 92.45 \\
\midrule
VLA-Cache (70\%) & 97.4 & 99.2 & 95.8 & 92.2 & 96.15 \\
VLA-Cache (50\%) & 97.4 & 99.8 & 96.0 & 90.2 & 95.85 \\
VLA-Cache (30\%) & 97.8 & 99.2 & 95.4 & 89.2 & 95.40 \\
\midrule
\method{} (70\%) & 97.8 & 98.4 & 97.8 & 94.6 & \textbf{97.15} \\
\method{} (50\%) & 96.0 & 98.4 & 96.6 & 93.6 & \textbf{96.15} \\
\method{} (30\%) & 96.2 & 98.2 & 96.6 & 92.2 & \textbf{95.80} \\
\bottomrule
\end{tabular}
\end{table}

Table~\ref{tab:hardware} compares end-to-end latency and visual-prefix KV storage.
The visual-path baselines, including token pruning and visual KV caching, achieve at most a 1.09$\times$ speedup because they optimize only the visual path while leaving the dominant action-denoising cost unchanged.
\method addresses both sources of overhead: at 30\% retention, median latency decreases from 306.19 to 143.45\,ms (2.134$\times$ speedup) and P95 latency from 313.89 to 147.51\,ms (2.128$\times$ speedup), while the live prefix KV state decreases from 17.02 to 6.19\,MiB (2.75$\times$ reduction).
Taken together with the success results in Table~\ref{tab:success}, \method is the only evaluated method that maintains at least 95.8\% average success across all matched budgets while achieving over 2$\times$ end-to-end speedup.

\begin{table}[t]
\caption{End-to-end latency and visual-prefix storage on A100 over 40 tasks $\times$ 5 repeats. P50/P95 are in ms, speedup is computed from P50, and KV is in MiB.}
\label{tab:hardware}
\centering
\footnotesize
\begin{tabular*}{0.8\columnwidth}{@{\extracolsep{\fill}}lrrrr@{}}
\toprule
Method (ratio) & P50 & P95 & Spd. & KV \\
\midrule
Dense $\pi_{0.5}$ (100\%) & 306.19 & 313.89 & 1.000 & 17.02 \\
\midrule
FastV (70\%) & 285.02 & 290.98 & 1.074 & 9.81 \\
FastV (50\%) & 283.53 & 288.89 & 1.079 & 8.02 \\
FastV (30\%) & 280.54 & 286.59 & 1.091 & 6.22 \\
\midrule
VLA-ADP (70\%) & 285.47 & 291.03 & 1.073 & 11.43 \\
VLA-ADP (50\%) & 285.45 & 298.76 & 1.073 & 10.72 \\
VLA-ADP (30\%) & 283.05 & 290.03 & 1.082 & 10.00 \\
\midrule
VLA-Cache (70\%) & 288.85 & 295.55 & 1.060 & 12.52 \\
VLA-Cache (50\%) & 284.31 & 290.11 & 1.077 & 12.52 \\
VLA-Cache (30\%) & 283.51 & 293.79 & 1.080 & 12.52 \\
\midrule
\method{} (70\%) & 144.42 & 148.35 & 2.120 & 9.81 \\
\method{} (50\%) & 144.36 & 148.00 & 2.121 & 8.02 \\
\method{} (30\%) & \textbf{143.45} & \textbf{147.51} & \textbf{2.134} & \textbf{6.19} \\
\bottomrule
\end{tabular*}
\end{table}

\subsection{Component Ablation}
Fig.~\ref{fig:selection-visualization} visualizes the roles of the two spatial components. Compared with current-frame pruning, DATP preserves the predicted future interaction region, while LACT restructures the retained tokens into a contiguous mask that is easier to pack. 
Table~\ref{tab:fcar-ablation} separates the temporal and spatial contributions.
Rows 1--4 run on the full dense visual prefix with no visual pruning, varying only the action-reuse policy.
The no-reuse row reproduces dense inference exactly (306.19\,ms, 96.15\%), validating the measurement pipeline.
Both fixed-refresh configurations reduce latency relative to dense inference, but their success rates decrease to 95.15\% and 94.50\%, respectively, against 96.15\% for dense inference.
FCAR alone refreshes 27.5\% of the action-expert blocks on average yet reaches 96.10\% success at 152.43\,ms (2.01$\times$). Unlike a fixed schedule that applies the same reuse pattern to every query, FCAR uses task-conditioned motion score to adjust the refresh budget, allowing more reuse under relatively stable observations and more recomputation when the observations change. This adaptive allocation helps preserve dense-level success while reducing action-expert computation. Adding flow-guided visual pruning and physical packing at 70\% retention further reduces latency by 8.0\,ms to 144.42\,ms and increases success to 97.15\%. Together, these results show that the spatial and temporal components provide complementary gains.

\begin{table}[t]
\caption{Action-reuse and visual-pruning ablation on LIBERO. Ref.\ denotes the refreshed action-expert block ratio; P50/P95 report policy latency (ms).}
\label{tab:fcar-ablation}
\centering
\scriptsize
\begin{tabular*}{\columnwidth}{@{\extracolsep{\fill}}lrrrr@{}}
\toprule
Configuration & Ref. (\%) & Succ. (\%) & P50 & P95 \\
\midrule
No reuse ($B_t\!=\!10$) & 100.0 & 96.15 & 306.19 & 313.89 \\
Fixed ($B_t\!=\!5$) & 50.0 & 95.15 & 180.73 & 185.75 \\
Fixed ($B_t\!=\!3$) & 30.0 & 94.50 & 158.49 & 162.57 \\
FCAR (caching only) & 27.5 & 96.10 & 152.43 & 158.76 \\
Full \method{} (FCAR + pruning) & 27.5 & \textbf{97.15} & \textbf{144.42} & \textbf{148.35} \\
\bottomrule
\end{tabular*}
\end{table}

\begin{figure}[t]
    \centering
    \includegraphics[width=0.85\columnwidth]{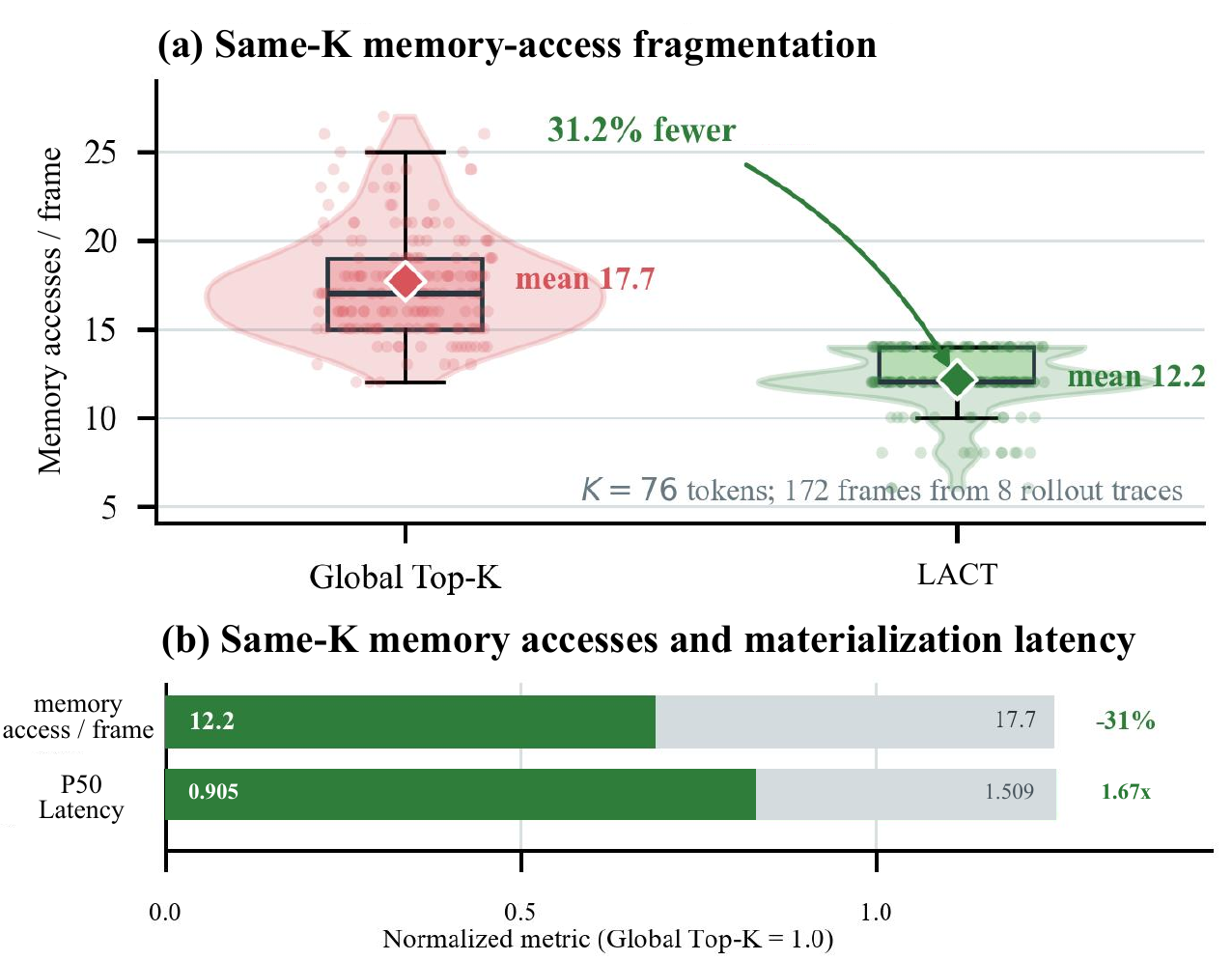}
    \caption{Same-budget packing-gather memory accesses and prefix-materialization latency of Global Top-$K$ and LACT.}
    \label{fig:hact-fragmentation}
    \vspace{-2mm}
\end{figure}

To compare memory-access efficiency at the same prefix length, Fig.~\ref{fig:hact-fragmentation} compares LACT with Global Top-$K$ over 172 frames from eight real rollout traces using the same $K=76$ token budget.
Across the 172 observation frames, LACT reduces the mean number of packing-gather memory accesses per frame from 17.7 to 12.2 (31.2\% fewer).
This more regular access pattern reduces prefix-materialization time from 1.509\,ms to 0.905\,ms per frame, corresponding to a 1.67$\times$ speedup relative to the fragmented Global Top-$K$ layout (Fig.~\ref{fig:hact-fragmentation}(b)).
Under the matched budget, both selectors produce the same compact KV capacity, so the measured materialization-speed difference is attributable to the memory-access pattern rather than prefix length.

\vspace{-0.6cm}

\FloatBarrier
\section{Conclusion}
We presented \method, a training-free framework that uses task-aware optical flow to jointly optimize visual token selection and action-expert refresh for VLA models.
DATP predicts future-critical tokens via motion propagation, LACT enforces contiguous masks for physical packing, and FCAR adapts refresh to causal motion dynamics.
On LIBERO, \method achieves 2.13$\times$ speedup and 2.75$\times$ KV reduction while preserving 95.80\% success at 30\% retention.

\bibliographystyle{IEEEtran}
\bibliography{reference_arxiv}

\end{document}